\documentclass{tADR2e}

\usepackage[hyphens]{url} 

\usepackage{comment}
\usepackage{enumerate}

\DeclareMathOperator*{\argmax}{argmax}

\allowdisplaybreaks

\usepackage{color}    
\usepackage{umoline} 
\usepackage{proofread} %
\noproofreadmark 

\begin{document}

\jvol{}
\jnum{}
\jyear{}
\jmonth{}

\title{
Spatial Concept-Based Navigation with Human Speech Instructions \\via Probabilistic Inference on Bayesian Generative Model
}

\author{
    Akira Taniguchi$^{a}$$^{\ast}$\thanks{$^\ast$Corresponding author. Email: a.taniguchi@em.ci.ritsumei.ac.jp\vspace{6pt}},
    Yoshinobu Hagiwara$^{a}$, Tadahiro Taniguchi$^{a}$, and Tetsunari Inamura$^{b}$ \\
    \vspace{6pt}
    $^{a}${\em{College of Information Science and Engineering, Ritsumeikan University, Shiga, Japan}};\\
    $^{b}${\em{National Institute of Informatics / The Graduate University for Advanced Studies, SOKENDAI, Tokyo, Japan}}\\
    \vspace{6pt}
    \received{v2.0 released July 2020} 
}

\maketitle

\begin{abstract}
Robots are required to not only learn spatial concepts autonomously but also utilize such knowledge for various tasks in a domestic environment.
Spatial concept represents a multimodal place category acquired from the robot's spatial experience including vision, speech-language, and self-position.
The aim of this study is to enable a mobile robot to perform navigational tasks with human speech instructions, such as `{\it Go to the kitchen}', via probabilistic inference on a Bayesian generative model using spatial concepts. 
Specifically, path planning was formalized as the maximization of probabilistic distribution on the path-trajectory under speech instruction, based on a control-as-inference framework. 
Furthermore, we described the relationship between probabilistic inference based on the Bayesian generative model and control problem including reinforcement learning.
We demonstrated path planning based on human instruction using acquired spatial concepts to verify the usefulness of the proposed approach in the simulator and in real environments.
Experimentally, places instructed by the user's speech commands showed high probability values, and the trajectory toward the target place was correctly estimated.
Our approach, based on probabilistic inference concerning decision-making, can lead to further improvement in robot autonomy.

    \medskip
\end{abstract}

\section{Introduction}
\label{sec:introduction}

\begin{figure}[tb]
  \begin{center}
    \includegraphics[width=0.90\hsize]{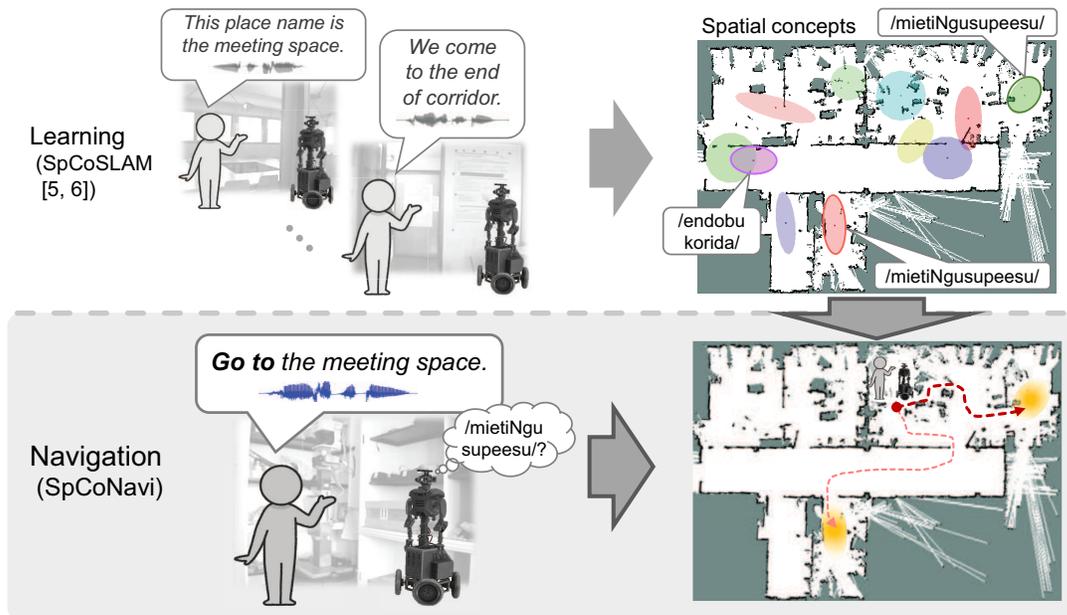}
    \caption{Overview of navigational tasks with human speech instruction. 
    Top: Learning phase for multimodal place categorization based on unsupervised learning in SpCoSLAM~\cite{ataniguchi_IROS2017,ataniguchi2020spcoslam2}.
    (i) The user teaches the robot the place name by an uttered sentence, while moving together. 
    (ii) The robot obtains the present estimated position, scenery, and speech signal, and acquires spatial knowledge, such as the relationship between words and places.
    Bottom: Navigation phase for path planning with learned spatial concepts and human speech instruction in SpCoNavi (proposed method).
    (iii) The user instructs the target place, such as `{\it Go to the meeting space.}'
    We assume specific trigger words for the execution of navigation, such as `{\it Go to}'.
    (iv) The robot estimates the path-trajectory toward a better target place using spatial concepts, as shown by the dotted bold red line. 
    In this case, the path to the closest of the two target places is selected.
    } 
    \label{fig:outline}
  \end{center}
\end{figure}

It is an important challenge for autonomous robots to perform various service tasks using concepts and vocabulary related to places that have been learned adaptively only from interaction with humans and environments~\cite{taniguchi2015symbol,taniguchi2019langrobo}.
This direction is required for robots to coexist with humans and operate in diverse living environments.
Particularly, navigational tasks from speech input could reduce the user's burden in a domestic environment.
To perform tasks through human-robot speech interaction, semantic mapping, a research field that deals with the meaning of places and grounding of spatial language in robots, is highly useful and effective~\cite{kostavelis2015semantic,landsiedel2017review}.
As a semantic mapping method including place categorization, 
a Bayesian generative model named SpCoSLAM that acquired spatial concepts, i.e., categorical knowledge of places from multimodal information through unsupervised learning, has been proposed~\cite{ataniguchi_IROS2017,ataniguchi2020spcoslam2}.
In this paper, we focus on a method by which a robot utilized spatial concepts learned from its own experience for navigational tasks. 
Path planning will be performed for the target state of a spatial concept conveyed through a human speech instruction, such as `{\it Guide to the kitchen.}' and `{\it Go to Daddy's room.}'. 
Our approach has the advantage that local names, which are learned without annotations, can be used for navigation in the individual environment for each family or community.
Therefore, users only need to speak to the robot from learning to task execution, namely they are not required to possess expertise in robotics.
Figure~\ref{fig:outline} shows the overview of learning and navigational tasks with human speech instruction. 

Here, to achieve path planning and navigation, decision-making is formulated based on a framework of control-as-inference (CaI), which is recently attracted attention~\cite{toussaint2009robot,levine2018reinforcement}.
The CaI bridged the theoretical gap between the probabilistic inference and the control problems including reinforcement learning (RL)~\cite{sutton1998reinforcement}. 
Mukadam et al. proposed the method for simultaneous trajectory estimation and planning via probabilistic inference~\cite{mukadam2017simultaneous,mukadam2019steap}. 
{
In other related approaches, the adaptation to human behavior via probabilistic inference in the navigation for social robots~\cite{kretzschmar2016socially} and the integration of RL and imitation learning based on the CaI framework~\cite{kinose2020integration} has been explored.
}
Generally, RL is used for planning and decision-making, including maze-solving tasks by robots.
However, in indoor-environment navigational tasks, there is a problem that it is not realistic to learn a policy by trial-and-error methods for RL each time the target is displaced by human instructions. 
Further, in general RL, task-dependent knowledge tends to be learned.
Instead, spatial concepts have the advantage that can be applied to various tasks because it is task-independent knowledge.
Therefore, our method adopts a probabilistic inference based on the CaI framework for the generative model of SpCoSLAM.

Path planning estimates a path-trajectory from the current position to the goal; however, it is not practical to explicitly obtain the robot with a definitive goal-point each time.
Generally, conventional global path planning methods, e.g., the A$^{\star}$ search algorithm, probabilistic road-map (PRM)~\cite{svestka1996probabilistic}, rapidly-exploring random trees (RRT)~\cite{lavalle1998rapidly}, as well as extension methods~\cite{karaman2011sampling}, need specify $\mathrm{xy}$-coordinates of target-posture or goal-point on the map.
There is a limitation in specifying these coordinates in that they depend strongly on the goal-point determination method. 
Alternatively, recent research on navigational tasks have used vision and language as signals to move toward the target~\cite{kollar2010toward, nair2018visual, fu2018from, anderson2018vision, noguchi2018navigation, fang2019smt}. 
Our method can autonomously determine actions toward the target place from only human speech instructions, using spatial concepts, which are the internal spatial knowledge formed bottom-up.

Previous studies~\cite{ataniguchi_IROS2017,lotfi2018WRS} did not discuss theoretical methods of applying spatial concepts to decision-making for path-trajectory.
As a substitute, these studies estimated the likeliest goal-position from a speech signal, using the spatial concepts. 
This approach is adopted in place recognition experiments using a speech signal in \cite{ataniguchi_IROS2017}.
Additionally, a similar heuristic method was adopted in robotics competitions for designing life-support in domestic environments, e.g., RoboCup@Home league and World Robot Summit 2018~\cite{lotfi2018WRS}. 
However, a conventional approach using spatial concepts has a problem in that a goal for an inappropriate place far from the current position may be determined.
To solve these issues, we realize navigation that considers the optimality of the entire trajectory theoretically on the CaI framework. 

The objective of this study is to enable a mobile robot to use spatial concepts to perform navigational tasks through an approach integrating the control problems represented by RL into the probabilistic inference on a Bayesian generative model mathematically.
Accordingly, this study focuses on action determination for a path-trajectory by robots on model parameters learned by SpCoSLAM. 
Moreover, we conducted a navigational task experiment based on human-robot interactions using acquired spatial concepts to verify the usefulness of the proposed approach. 
Our approach has the following advantages. 
The robot can reach the closest and most suitable place among several places under an ambiguous instruction, e.g., `{\it Go to bedroom}' in three-bedroom home environments.
Additionally, the robot can successfully navigate through various instructions using a many-to-many correspondence between places and words, 
e.g., cases where (i) multiple places called by the same name exist, (ii) multiple names in the same place exist, and (iii) the target place is specified by a speech instruction containing multiple place names.

The main contributions of our study are as follows:
\begin{itemize}
 \item       We realized global path planning with speech instruction under spatial concepts, acquired from the bottom-up by the robot, without setting an explicit goal-position. 
 \item       We formulated decision-making as the maximization of trajectory probability based on the CaI framework, with the same generation process as in SpCoSLAM. 
\end{itemize}

The remainder of this paper is organized as follows.
In Section~\ref{sec:SpCoSLAM}, we present an overview of the Bayesian generative model for learning, along with the formulation and online learning algorithm, SpCoSLAM.
In Section~\ref{sec:spconavi}, we present our method SpCoNavi by probabilistic inference for navigation and path planning.
In Section~\ref{sec:experiment1}, we demonstrate the experimental result by evaluation and comparison in the simulator environments.
In Section~\ref{sec:experiment1-2:example}, we demonstrate the application potentiality of SpCoNavi through some examples in the simulator environments.
In Section~\ref{sec:experiment2}, we discuss the effectiveness of our method in a real environment.
Section~\ref{sec:conclusion} concludes the paper.

\section{Background: Online learning for spatial concepts and lexical acquisition with mapping}
\label{sec:SpCoSLAM}
We introduce the outline of SpCoSLAM~\cite{ataniguchi_IROS2017} as an unsupervised learning method for acquisition of spatial knowledge to use in a navigational task.
{
Our study is specific to the validation of navigation tasks, and more detailed information regarding learning algorithms has been presented in previous reports~\cite{ataniguchi_IROS2017,ataniguchi2020spcoslam2}.
}

\begin{figure}
    \begin{center}
        \includegraphics[width=160pt]{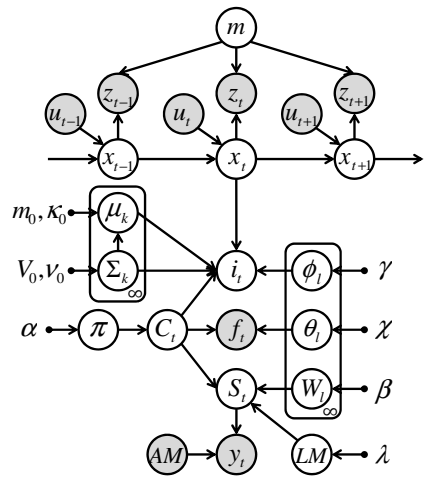} 
        \caption{
            Graphical model representation of SpCoSLAM for learning \cite{ataniguchi_IROS2017}. 
            The graphical model represents the conditional dependency between random variables.
            Gray nodes indicate observation variables, and white nodes denote unobserved latent variables.
            SpCoSLAM, an integrated model, can perform multimodal spatial concept formation, lexical acquisition, and SLAM simultaneously. 
        }
        \label{fig:graphical-model}
    \end{center}
\end{figure}

\begin{table}
    \tbl{Description of the random variables used in the graphical model}{
        \begin{tabular}{cl} 
            \toprule
            \textbf{Symbol} & \textbf{Definition}\\
            \colrule
$m$ & Environmental map \\ 
$x_t$ & Self-position of a robot (state variable) \\ 
$u_t$ & Control data (action variable) \\ 
$z_t$ & Depth sensor data \\ 
$f_{t}$ & Visual feature of a camera image \\ 
$y_{t}$ & Speech signal of an uttered sentence \\ 
$i_{t}$ & Index of position distributions \\ 
$C_{t}$ & Index of spatial concepts \\ 
{$S_t$} & {Word sequence (recognition result in an uttered sentence)} \\ 
{$\mu_{k}$, $\Sigma_{k}$} & {Parameters of multivariate normal distribution (position distribution)} \\ 
{$\pi $} & {Parameter of multinomial distribution for index $C_t$ of spatial concepts} \\ 
{$\phi_{l}$} & {Parameter of multinomial distribution for index $i_{t}$ of the position distribution} \\ 
{$\theta_{l}$} & {Parameter of multinomial distribution for the image feature} \\ 
{$W_{l}$} & {Parameter of multinomial distribution for the names of places} \\ 
$LM$ & Language model (n-gram and word dictionary) \\ 
$AM$ & Acoustic model for speech recognition \\ 
$\alpha $, $\beta$, $\gamma$, $\chi$, $\lambda$, & {Hyperparameters of prior distributions} \\
$m_{0}$, $\kappa_{0}$, $V_{0}$, $\nu_{0}$& \\
            \botrule
        \end{tabular}
    }
    \label{tab:graphical-model}
\end{table}

\subsection{Overview}
\label{sec:SpCoSLAM:overview}
SpCoSLAM integrates simultaneous localization and mapping (SLAM)~\cite{thrun2005probabilistic} with a place categorization based on multimodal information including positions, speech-language, and image features.
Figure~\ref{fig:graphical-model} shows the graphical model representation of SpCoSLAM for learning the global parameters, and Table~\ref{tab:graphical-model} lists each variable of the graphical model. 
The details of the formulation of the generative process represented by the graphical model are described in Appendix~\ref{apdx:SpCoSLAM:overview:formulation}.

The online learning algorithm in SpCoSLAM has the advantage that can simultaneously learn a map, spatial concepts, and a language model without any prior knowledge of the environment or place names. 
Further, it learns the many-to-many correspondences between places and words via spatial concepts and can mutually complement the uncertainty of information by using multimodal information.
Via this method, the robot learns sequential spatial concepts for unknown environments without maps. 
Lexical acquisition, including acquisition of unknown words, is also possible via sequential updating of the language model.
Further, this method can estimate an appropriate number of clusters of spatial concepts and position distributions depending on the data using the so-called online Chinese restaurant process~\cite{aldous1985exchangeability}.
Additionally, to enable long-term learning for a real robot with limited calculation resources, a scalable and improved online learning algorithm was developed~\cite{ataniguchi2020spcoslam2}.

\subsection{Formulation for the online learning algorithm}
\label{sec:SpCoSLAM:learning}
The online learning algorithm introduces sequential equation updates to estimate the parameters of the spatial concepts into the formulation of a Rao-Blackwellized particle filter~\cite{doucet2000rao} in the FastSLAM~2.0~\cite{montemerlo2003fastslam} and its grid-based SLAM~\cite{gridbasedfastslam2007}.
The particle filter is advantageous in that parallel processing can be easily applied because particles can be calculated independently.
{Theoretically, other particle-filter-based SLAMs besides FastSLAM 2.0 can also be used.}

In the formulation of SpCoSLAM, the joint posterior distribution can be factorized to the probability distributions of a language model $LM$, a map $m$, the set of model parameters of spatial concepts $\Theta = \{ {\mathbf W}, \mbox{\boldmath $\mu $}, \mbox{\boldmath $\Sigma$}, {\mathbf \theta}, {\mathbf \phi}, \pi \}$, the joint distribution of the self-positions $x_{0:t}$, and the set of latent variables $\mathbf{C}_{1:t} = \{i_{1:t},C_{1:t},S_{1:t} \}$.
We describe the joint posterior distribution as 
\begin{eqnarray}
&&p(x_{0:t},\mathbf{C}_{1:t}, LM, \Theta, m 
\mid u_{1:t}, z_{1:t}, y_{1:t}, f_{1:t}, AM ,\mathbf{h}) \nonumber \\
&&=p(LM \mid S_{1:t}, \lambda)
p(\Theta \mid x_{0:t}, \mathbf{C}_{1:t}, f_{1:t}, \mathbf{h})p(m \mid x_{0:t}, z_{1:t}) \nonumber \\
&&\hspace{1.0em}\cdot~\underbrace{p(x_{0:t},\mathbf{C}_{1:t} \mid u_{1:t}, z_{1:t}, y_{1:t}, f_{1:t}, AM ,\mathbf{h})}_\text{Particle~filter}
\label{eq:spcoslam}
\end{eqnarray}
where the set of hyperparameters is denoted by $\mathbf{h}= \{ \alpha,\beta,\gamma,\chi,\lambda, m_{0},\kappa_{0}, V_{0},\nu_{0} \}$.

The learning procedure for each step is described in Appendix~\ref{apdx:SpCoSLAM:learning:procedure}.
The variables of the joint posterior distribution can be learned by Gibbs sampling, which is a Markov chain Monte-Carlo-based batch learning algorithm, in a manner similar to the nonparametric Bayesian spatial concept acquisition method (SpCoA)~\cite{taniguchi_spcoa}.
{In addition, it is also possible to perform learning in a spatial concept formation model after the map is generated via any other SLAM.}

\section{SpCoNavi: Spatial Concept-Based Navigation under Human Speech Instructions}
\label{sec:spconavi}
We propose a spatial concept-based navigation method (SpCoNavi) from human speech instructions, using probabilistic inference on the graphical model in the generative process in the same manner as in SpCoSLAM.
Here, we define the navigation and path planning problem as trajectory probability maximization under human speech instruction. 
This problem definition is different from the conventional setting that the robot extracts the goal-position explicitly from human speech and navigates toward it.

In the probabilistic inference, various uncertainties occurring in speech recognition and self-localization need be addressed. 
Our task setting is an extended version of a partially observable Markov decision process (POMDP) with several uncertainties because SpCoSLAM is a model that integrates multimodal categorization into Bayesian filter-based SLAM. 
Moreover, it finds an inference based on an overall integrated model, allowing it to make complementary inferences of multimodal information that consider uncertainties in random variables.
We consider this to enable overall optimal path planning in the model.

\begin{figure}[tb]
        \begin{center}
            \includegraphics[width=210pt]{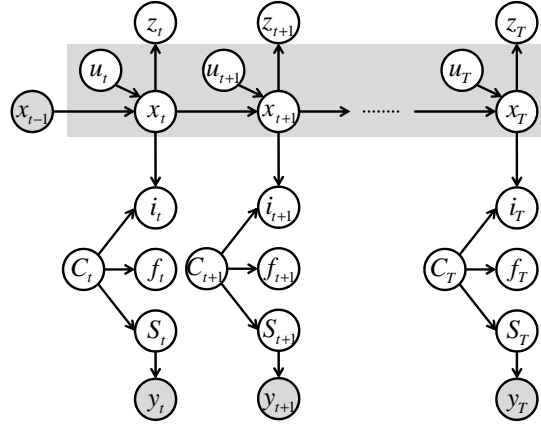}
        \end{center}
    \caption{
        Graphical model representation of the proposed SpCoNavi for navigation and path planning.
        The global parameters are omitted from the SpCoNavi graphical model.
        Variables within a grayed rectangular region are the estimated targets in our method. 
    } 
    \label{fig:gmodel_navigation_simple}
\end{figure}

\subsection{Formulation and derivation}
\label{sec:spconavi:formulation}
A reasonably accurate map of the environment, spatial concepts, and lexicon, are assumed to have already been acquired. 
Consequently, SpCoSLAM has been in operation in the environment and each model parameter for the learning result from the last time-step has been fixed. 
Figure~\ref{fig:gmodel_navigation_simple} shows a graphical model for path planning for navigation based on SpCoSLAM. 
The subscript $T$ denotes the last time-step of path planning, and when the start time-step is $t=1$, it denotes a planning horizon. 
Here, the mathematical formula is denoted as $t=1$. 

Our method estimates an action sequence $u_{1:T}$ (and the path $x_{1:T}$ on the map) in which the probabilistic distribution representing trajectory $\tau=\{ u_{1:T}, x_{1:T} \}$ is maximized when a human speech instruction $y_{t}$ is given. 
The equation used to identify the action sequence when the probability distribution that is the objective function takes the maximum value is as
\begin{eqnarray}
u_{1:T} &=& \argmax_{u_{1:T}} p(\tau \mid y_{1:T}, x_{0}, \Theta_{G}). 
\label{eq:spconavi}
\end{eqnarray}
A set of learned global parameters include the map $m$, a set of model parameters represent spatial concepts $\Theta$, the acoustic model $AM$, and the language model $LM$ is denoted as $\Theta_{G} = \{ m, \mathbf{W}, \mbox{\boldmath $\mu $}, \mbox{\boldmath $\Sigma$}, \mathbf{\theta}, \mathbf{\phi}, \pi, AM, LM \}$.
Here, we assume that the self-position $x_{0}$ at a previous time-step is provided. 
In reality, the default position set beforehand or an estimated value obtained through a self-localization method such as Monte-Carlo localization (MCL)~\cite{dellaert1999monte} is available.

{
As shown in Equation~(\ref{eq:spconavi}), an optimal trajectory along which to travel to the position indicated by the spatial concept is estimated by maximizing the posterior probability distribution conditioned on the quasi-observation $y_{1:T}$. 
Here, speech instruction $y_{t}$ is an observed variable denoting an optimal goal-signal. 
We assume that the same observations are continuously noted from the current time-step, $t=1$, to the planning horizon $T$ as quasi-observation $y_{1:T}$. 
This assumption is based on the fact that an observation $y_{t}$ corresponds to a binary random variable ${\cal{O}}_{t}$ that indicates whether a state-action pair is optimal in the CaI framework~\cite{levine2018reinforcement} (see Section~\ref{sec:spconavi:capi} and Appendix~\ref{apdx:SpCoNavi:CaI}). 
From a different perspective, this assumption can be interpreted as the robot internally repeating the speech instruction content for a certain time period; i.e., it remains conscious to arrive at the destination.
}

To calculate Equation~(\ref{eq:spconavi}), the latent variables, depth sensor data $z_{1:T}$, indices of position distribution $i_{1:T}$, indices of spatial concept $C_{1:T}$, and word sequences $S_{1:T}$ must be marginalized.
Subsequently, although image feature $f_{1:T}$ is also an unobserved variable, it does not have to be considered in the calculation because of the conditional independence relationships among the variables in the graphical model.

Furthermore, the set of $N$-best speech recognition results $S_{T}^{N}$, which is represented by bag-of-words (BoW), is used 
because the direct calculation of the probability value representing speech recognition $p(S_{T} \mid y_{T}, AM, LM)$ is difficult.
This makes it possible to deal with uncertainty in speech recognition errors.
The speech instruction $y_{T}$ is assumed as the same observation from $t=1$ to $T$; therefore, the same BoW $S_{T}^{N}$ is copied for the entire time.

Moreover, we assume that $p(u_{t})$ and $p(x_{t})$ are uniform distributions. 
Additionally, because the term obtained through $z_{t}$ marginalization only affects a measurement model $p(z_{t} \mid x_{t}, m)$ for self-localization, it can be expressed as $\sum_{z_{t}} p(z_{t} \mid x_{t}, m)=1$.

\begin{figure}
    \begin{center}
        \includegraphics[width=0.70\textwidth]{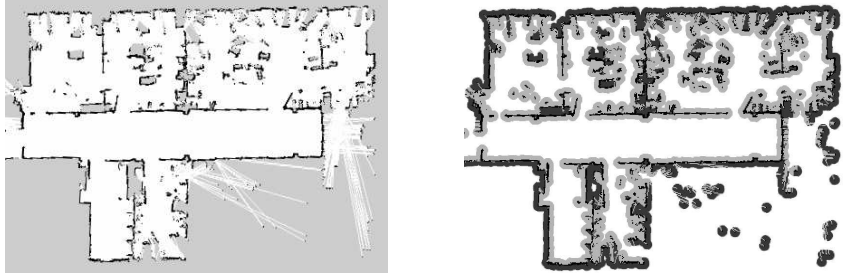}
        \caption{Example visualized in gray scale. Left: the occupancy grid map. Right: the cost map based on the grid map.}
        \label{fig:map_costmap}
    \end{center}
\end{figure}

To avoid obstacles on the map,
the map-based motion model~\cite{thrun2005probabilistic} would be more appropriate, as
$p(x_{t} \mid u_{t}, x_{t-1}, m) 
\propto p(x_{t} \mid m)p(x_{t} \mid u_{t}, x_{t-1})$.
We can utilize a motion model while considering the cost map $p(x_{t} \mid m)$, where the probability of an obstacle being in a cell is 0, and the closer the robot approaches the obstacle, the lower the probability is, i.e., a probabilistic expression of robot's configuration space (C-space).
Figure~\ref{fig:map_costmap} shows the visualization example of the grid map and the cost map in the environment.

Consequently, the trajectory posterior probability distribution given the condition ${y}_{t}$ at all times $t \in \{ 1, \dots, T\}$  is shown as 
\begin{eqnarray}
p(\tau \mid y_{1:T}, x_{0}, \Theta_{G})  
&\approx& \prod_{t=1}^{T} p(x_{t} \mid u_{t}, x_{t-1})  \prod_{t=1}^{T}  \Biggl[ p(x_{t} \mid m) 
\sum_{C_{t}} \Biggl[ {\rm Mult}(S_{T}^{N} \mid W_{C_{t}}){\rm Mult}( C_{t} \mid \pi) \Biggr.  \nonumber \\  
&&\hspace{0pt} \Biggl.\Biggl. \sum_{i_{t}} {\cal N}(x_{t} \mid \mu _{i_{t}}, \Sigma_{i_{t}}) \Biggr.{\rm Mult}(i_{t} \mid \phi_{C_{t}}) \Biggr] \Biggr], 
\label{eq:spconavi_marginalization2}
\end{eqnarray}
where ${\rm Mult}()$ is multinomial distribution and ${\cal N}()$ is multivariate normal distribution.

The probability value is calculated for each state in the environment, and the path-trajectory is estimated using a Viterbi algorithm based on dynamic programming.

\subsection{Correspondence relationship with RL and control as probabilistic inference}
\label{sec:spconavi:capi}

We discuss the relationship between SpCoNavi and RL from the perspective of CaI. 
The detail of the formulation of CaI is described in Appendix~\ref{apdx:SpCoNavi:CaI}.

In the graphical model of SpCoNavi, the probabilistic distribution that corresponds to the generative process on the optimality variable ${\cal{O}}_{t}$ in \cite{levine2018reinforcement} is represented as 
\begin{eqnarray}
p(y_{t} \mid x_{t}, \Theta_{G}) &=& \exp(r_{\Theta_{G} \mid y_{t}}(x_{t})).
\label{eq:spconavi_bainary}
\end{eqnarray}
Equation~(\ref{eq:spconavi_bainary}) implies that the continuous speech signal $y_{t}$ is an extension of the binary random variable ${\cal{O}}_{t}$ in the CaI framework~\cite{levine2018reinforcement} to a continuous, many-valued variable.
Note that the speech signal of the uttered sentence at the time-step $t$ is actually real-time time-series data, i.e., $y_{t} = ( y_{t,1},  y_{t,{2}}, \dots,  y_{t,{D_{t}}} )$.
Subsequently, $u_{t}$ has been eliminated from the dependency of the random variables based on the graphical model.

Equation~(\ref{eq:spconavi_marginalization2}) before marginalization can be expressed as 
\begin{eqnarray}
p(\tau \mid y_{1:T}, x_{0}, \Theta_{G})  
&\propto& \prod_{t=1}^{T} p(x_{t} \mid u_{t}, x_{t-1}) p(y_{t} \mid x_{t}, \Theta_{G}) \nonumber \\
&=& \left[ \prod_{t=1}^{T} p(x_{t} \mid u_{t}, x_{t-1}) \right] \exp \left( \sum_{t=1}^{T} r_{\Theta_{G} \mid y_{t}}(x_{t}) \right). 
\end{eqnarray}

Here, the reward function $r_{\Theta_{G} \mid y_{t}}(x_{t})$ parameterized by $\Theta_{G}$ under the condition $y_{t}$ is shown as 
\begin{eqnarray}
r_{\Theta_{G} \mid y_{t}}(x_{t}) &=& \log( p(y_{t} \mid x_{t}, \Theta_{G}) ).
\label{eq:spconavi_reward}
\end{eqnarray}
This reward function is nothing less than the log-likelihood of the emission probability in Equation~(\ref{eq:spconavi_marginalization2}).
In other words, changing the observation $y_{t}$ changes the destination, at which a higher reward is given.
Additionally, this approach enables the robot to perform navigational tasks without explicitly providing manually designed rewards, indicating that it can avoid problems associated with reward design in RL.

The policy function can be shown as
$\pi_{\vartheta}(x_{t}, u_{t+1}) = p(u_{t+1} \mid x_{t}, \vartheta)$, 
where $\vartheta$ is a parameter for the policy function.
The optimal policy function can be shown as 
\begin{eqnarray}
p(u_{t+1} \mid x_{t}, \vartheta^{\ast}) &\approx& p(u_{t+1} \mid x_{t}, {y}_{t:T}, \Theta_{G}),  
\label{eq:spconavi_optimalpolicy}
\end{eqnarray}
where $\vartheta^{\ast}$ is an optimal policy parameter. 
Equation~(\ref{eq:spconavi_optimalpolicy}) indicates that $\vartheta^{\ast}$ is a parameter for an approximate distribution in the probability distribution of the right side.
However, using our path planning method, Equation~(\ref{eq:spconavi_marginalization2}) can be inferred from the probability distributions defined in the generative process.
Therefore, policy searches such as that performed in RL can be substituted by learning model parameters $\Theta_{G}$.

\subsection{Approximate inference by A$^{\star}$ search algorithm}
\label{sec:spconavi:approx}

We discuss and propose an approximate inference method for speeding up using the A$^{\star}$ search algorithm.
In this case, the state $x_{t}$ and the action $u_{t}$ are discretized.
{
In the inference obtained via the Viterbi algorithm, the computational cost will increase according to the value of the planning horizon. 
If the size of the environmental map increases, i.e., the number of states, then the planning horizon should be set to a large value.
}

We can consider a method of considering the cost map and emission probability into the A* and selecting the closest one or the most suitable one among several candidates.

\begin{description}
\item[(a)] 
When the robot hears a user's speech signal $y_{t}$ including a place's name, it estimates the possible goal-positions.
The positions are estimated as follows:
\begin{eqnarray}
x_{t}^{(j)} &\sim& p(x_{t} \mid y_{t}, \Theta_{G}) 
\approx  p(x_{t} \mid S_{t}^{N}, \Theta_{G}), \quad S_{t}^{N} = \argmax_{S_{t}^{N}} p(S_{t} \mid y_{t}, AM, LM), 
\label{eq:xbest0_sampling}
\end{eqnarray}
where the $j$-th sampled position is denoted as $x_{t}^{(j)}$, and the $N$-best speech recognition result is denoted as $S_{T}^{N}$.
The number of possible positions is $J$.
Here, it is difficult to calculate Equation~(\ref{eq:xbest0_sampling}) for all possible positions.
Therefore, we use the mean value or the sampled value for each position distribution with the highest probability as a position candidate $x_{t}^{{(j)}}$.

\item[(b)] 
The robot sets the estimated position $x_{t}^{(j)}$ as the goal and plans a path using the A$^{\star}$ search algorithm.
The cost function of A$^{\star}$ uses the negative log-likelihood of the emission probability in Equation~(\ref{eq:spconavi_reward}).

\item[(c)] 
The robot selects a trajectory, from the start to the goal, that has the minimum cumulative cost among all candidates.

\end{description}

This method appears as a facile and heuristic extension of the conventional method of estimating a goal-point using spatial concepts.
Therefore, theoretical optimality is not guaranteed. 
The inference by the Viterbi algorithm can theoretically estimate the optimal solution within the CaI framework. 
However, we believe that this method using A$^{\star}$ is positioned as a greedy approximate inference method for our method. 
This approximate method loses the optimality on the entire trajectory; however, a solution close to the optimal solution or a sub-optimal solution may be obtained.

\section{Experiment I-A: Simulator environment}
\label{sec:experiment1}
We demonstrate that a robot with acquired spatial concepts can successfully plan a path using our method.
Moreover, we verify whether it can appropriately navigate to a nearby place when multiple places denoted by the same name exist in the environment.

\subsection{Condition}
\label{sec:experiment1:condition}
The simulator environment was SIGVerse version 3.0~\cite{inamura2020sigverse}, a client--server based architecture that can connect the robot operating system (ROS)~\cite{ros} and Unity.
The virtual robot model in SIGVerse was Toyota's Human Support Robot (HSR)~\cite{HSR2019}.
{
The diameter of the HSR is $\varphi$430 mm. 
The robot was equipped with a laser range sensor. The depth values, which were obtained using a laser range sensor, and pseudo odometry, which was simulated according to the control values of movement using a simulator, were used for mapping.
} 
We used 10 three-bedroom home environments\footnote{3D home environment models are available at \url{https://github.com/a-taniguchi/SweetHome3D_rooms}.}, with different layout and room size, created via Sweet Home 3D\footnote{Sweet Home 3D: \url{http://www.sweethome3d.com/}}, a free software for interior design.
The robot performed SLAM to generate an environment map using the {\tt gmapping} package in the ROS. 
We used the {\tt navigation} and {\tt costmap\_2d} package in the ROS to obtain a global cost map. 
The spatial concepts and position distributions were {9 or }10 for each environment.
For each place, on average, 15 training datasets were provided.
The latent variables $C_{t}$ and $i_{t}$ were assumed to be almost accurately estimated, and model parameters of the spatial concept were obtained via Gibbs sampling, similar to~\cite{taniguchi_spcoa}.
{The visual features of a camera-acquired image were not treated.} 
The word dictionary was provided in advance.
The hyperparameters for learning were set as follows: $\alpha=1.0$, $\gamma=1.0$, $\beta=0.1$, $\chi=0.1$, $m_{0}=[ 0 , 0 ]^{\rm T}$, $\kappa_{0}=0.001$, $V_{0}={\rm diag}(2,2)$, and $\nu_{0}=3$. 
The planning horizon was $T=200$.
The robot's initial position was set from arbitrary movable coordinates on the map. 
The user provided a word as an instruction of the target place's name.

Although the state of self-position $x_{t}$ takes a continuous value in reality, here, it is expressed discretely for each movable cell in the occupancy grid map $m$. 
The position represents the two-dimensional coordinate as $x_{t}= (\mathrm{x}, \mathrm{y})^{\mathrm{T}}$
The motion model---state transition probability---assumes a deterministic model. 
The control value $u_{t}$ is assumed to move one cell from its current position on the map per time-step.
The action $u_{t}$ is discretized into stay, forward, backward, and side-to-side movements, i.e., the set of action types is $\mathcal{A} = $ \{{\tt stay}, {\tt up}, {\tt down}, {\tt left}, {\tt right}\}.

\subsection{Evaluation metrics and comparison methods}
\label{sec:experiment1:evaluation}
For evaluation, we corrected the position within the rectangular area surrounding the position coordinates provided as the same name of place. 
The navigation success rate (NSR), a metric for evaluating whether a generated path reaches the indicated target place, is calculated as ${\rm NSR} =  {n_{\rm C}}/{n_{\rm U}}$, 
where $n_{\rm U}$ denotes the number of trials and $n_{\rm C}$ the number of correct positions.
Additionally, when several places have the same name, the rate at which the robot reached the closest place from the initial position is considered the Near-NSR.
Furthermore, we evaluate the estimated path-length (PL) when the robot reaches the target place.

We compare the performance of following methods: 
\begin{description}
\item[(A) SpCoNavi (Proposed method)]
This method uses the probabilistic inference based on the Viterbi algorithm, as shown in Section~\ref{sec:spconavi:formulation}.

\item[(B) SpCoNavi (Approximate inference)] 
This method uses an approximate inference based on the A$^{\star}$ search algorithm, as shown in Section~\ref{sec:spconavi:approx}.
The number of position candidates is $J=10$.
The heuristic function of A$^{\star}$ uses $-\log(p(u_{t}))$ times the Manhattan distance from the current position to the temporary goal-position.
The probability of an action is $p(u_{t})= 1/|\mathcal{A}|$, where $|\mathcal{A}|$ is the number of action types.

\item[(C) Baseline method (Spatial concept)]
This method uses spatial concepts as a conventional approach.
When the robot hears a user's speech signal $y_{t}$ including a place's name, it estimates a position $x_{t}^{\ast}$ indicated by the uttered sentence.
This estimation was calculated as $x_{t}^{\ast} = \argmax_{x_{t}} p(x_{t} \mid y_{t}, \Theta_{G})$.
It was approximated using the $N$-best speech recognition results $S_{t}^{N}$ in the same manner as~\cite{ataniguchi_IROS2017}. 
We use the mean values of the position distribution as position candidates for $x_{t}^{\ast}$.
The robot sets the estimated position $x_{t}^{\ast}$ as the goal and plans a path by using the A$^{\star}$ search algorithm.
The heuristic function of A$^{\star}$ uses the Manhattan distance from the current position to the goal.
The cost map value is used as the cost of a position to generate a path away from obstacles. 
The cost of an action is 1.0 for each time-step.

\item[(D) Baseline method (Database)]
This method uses database without spatial concept formation as a conventional approach.
Among the datasets used for learning the spatial concepts, a goal-position, which includes the word in the instruction, is randomly selected.
The robot sets the estimated position $x_{t}^{\ast}$ as the goal and plans a path by using the A$^{\star}$ search algorithm.
The setting of A$^{\star}$ is the same as (C).

\item[(E) Baseline method (Random)]
This method randomly selects the goal from all positions in the dataset, as chance-level.
The robot sets the selected position as the goal and plans a path using the A$^{\star}$ search algorithm.
The setting of A$^{\star}$ is the same as (C).
\end{description}

The experiment is a comparison between SpCoNavi (A) and the conventional methods (C, D), rather than a comparison with A$^{\star}$.
Therefore, the experimental comparison focuses on how to use spatial concepts for path planning.
More generally, these baselines correspond to the classical approaches of identifying only the goal-position from any semantic map in other studies.

\subsection{Result of the comparison of methods}
\label{sec:experiment1:result}

\begin{table}[tb]
\begin{center}
\caption{Evaluation result in simulator environment{: The underlined numbers in bold indicate the highest evaluation values, and the underlined numbers indicate the second-highest evaluation values.}}
\begin{tabular}{lccc} \hline \noalign{\smallskip} 
\textbf{Methods}                       & \textbf{NSR}  & \textbf{Near-NSR} & \textbf{PL} \\ 
\noalign{\smallskip}\hline\noalign{\smallskip} 
(A) SpCoNavi                   & \underline{\textbf{1.00}} & \underline{\textbf{0.75}}     & \underline{63.15}       \\ 
(B) SpCoNavi (Approx.)         & 0.85 & \underline{0.60}     & \underline{\textbf{48.88}}        \\ 
(C) Baseline (Spatial concept) & \underline{0.90} & 0.35     & 84.11       \\
(D) Baseline (Database)        & 0.95 & 0.30     & 88.26       \\
(E) Baseline (Random)          & 0.35 & 0.10     & 106.14      \\
\noalign{\smallskip} \hline 
\end{tabular}
\label{table:result}
\end{center}
\end{table}

Table~\ref{table:result} shows the average values of 20 trials of NSR, Near-NSR, and PL when the word `{\it bedroom}' was provided as an instruction.
Compared to baseline methods, ours showed higher accuracy and a shorter path-length.
Particularly, our method could estimate the trajectory from the current position to the closest place when multiple places with the same name existed.

The NSR of baselines (C, D) did not reach 1.00 because of the goal-coordinate estimation method, not A$^{\star}$. 
Baselines have a risk that the goal-coordinates may be set to a non-movable point with an obstacle.
In path planning failure in (C), there was a case where the center of the formed spatial concept was in a no-free-space cell position, and the goal was set at that position.
In (D), there was a case where a goal was set on the obstacle or the outside of the room because of the self-localization error and the displacement of the map.
However, our method could estimate the path toward the movable cell by considering the joint probability on the trajectory.
Furthermore, the baseline (C) could not estimate a path toward a nearby place because it selected the position representing the most instructed place name regardless of the initial value.

\begin{figure}[tb]
  \begin{center}
        \includegraphics[clip, width=\hsize]{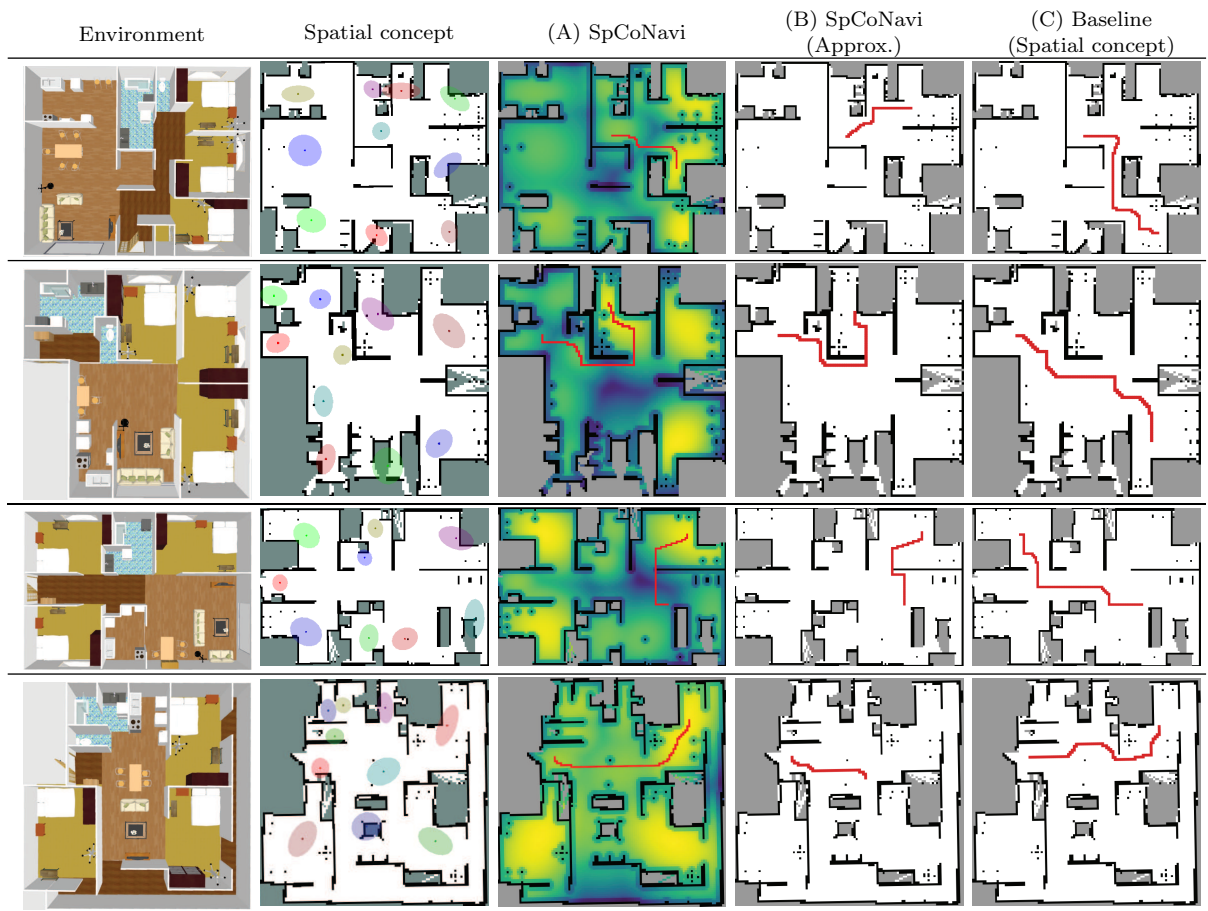}    
    \caption{
        Path planning results for each method using spatial concepts. 
        From the left, these are arranged in the order of the overhead view of the home environments on SIGVerse, position distributions of learned spatial concepts on the map, and emission probabilities (color maps on log scale) and path trajectories (red) for each method. 
        Ellipses in the spatial concept have drawn position distributions that are multivariate normal distributions.
        The color of each ellipse in the spatial concept was randomly determined for each environment.
        The color map has a high value for bright yellow and a low value for dark purple.
    }
    \label{fig:result_SIGVerse}
  \end{center}
\end{figure}

Figure~\ref{fig:result_SIGVerse} shows the example of the environment, learned spatial concepts, and path planning results obtained by (A -- C), respectively. 
Results of baseline (D, E) were omitted from the figure because these were the same as or worse than (C).
In our method, the emission probability on a log scale---the log-likelihood for each cell on the map---is expressed as a color map.
The estimated path-trajectory is denoted by a red line.
The initial position is different for each environment; however, it is the same for all methods in a given environment.
In our method, the emission probabilities of the bedrooms showed higher values than that of other places and the trajectory was estimated towards that place. 
Although A$^{\star}$ used the cost map as a cost function, our method could generate safe trajectories avoiding the obstacles by the probabilistic distribution representation considering the cost map derived from the entire internal structure of a Bayesian generative model.
Overall, our method showed its tendency to go towards the closest bedroom, even if the command was ambiguous, such as `{\it Go to bedroom}'.
Therefore, we could estimate the appropriate trajectory to reach the range of the target place contrary to conventional methods.

In the approximate inference (B), it tended to select the shortest path among candidates.
Therefore, the PL of (B) was smaller than (A).
In some cases, the path was too short to reach the target place, as shown in the last environment of Fig.~\ref{fig:result_SIGVerse}.
This result may be the limit of the heuristic extension of the conventional method using the spatial concept.
Compared to (A), the calculation time of (B) was reduced by 147.25 times on average.
{
The calculation times for (A) and (B) are 39.56$\times 10^{2}$ s and 26.86 s, respectively. 
The CPU was an Intel Core i7-6850K with 16GB DDR4 2133MHz SDRAM.
In the experiment, only one CPU core was used without parallel computation.
}

\begin{figure}
    \begin{center}
        \includegraphics[width=0.99\textwidth]{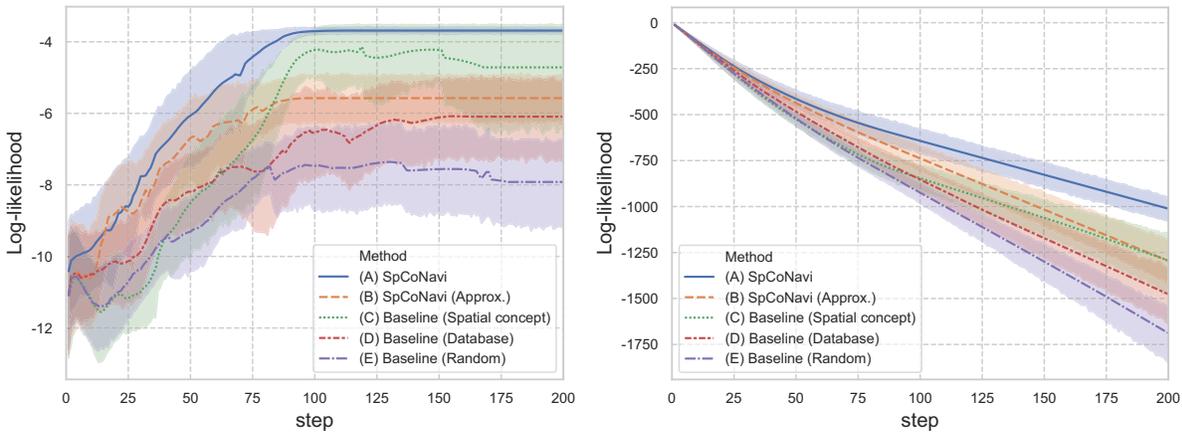}
        \caption{The log-likelihood (left: value in each step, right: cumulative value) for each time-step}
        \label{fig:log-likelihood}
    \end{center}
\end{figure}

Figure~\ref{fig:log-likelihood} shows the log-likelihood values per step and cumulative values up to the planning horizon.
If the estimated path is shorter than the planning horizon $T$, we assume that the robot will stay in end position until time-step $T$. 
SpCoNavi (A) had the highest log-likelihood on the whole.
Particularly, (A), which estimates the entire path-trajectory, obtained a higher likelihood in a shorter step than (C).
Consequently, our method can estimate a path with a high cumulative reward.
In Fig.~\ref{fig:log-likelihood}, (B) showed that tends to a log-likelihood value closer to (A) than other methods during the early steps in value per step (left figure), and all steps in cumulative value (right figure). 
It indicates that the approximation accuracy of (B) is high.
However, after step 80, the log-likelihood values per step of (C) were higher than those of (B) because the likeliest position in the entire environment was set as the goal even if it was far away in (C).

\section{Experiment I-B: Effective application examples using spatial concepts in simulator environment}
\label{sec:experiment1-2:example}
Here, we present the application potentiality of SpCoNavi in navigational task with human instructions.
We demonstrate qualitatively typical examples of navigational tasks according to our method when using a many-to-many correspondence between places and words.

\subsection{Situation setting and condition}
\label{sec:experiment1-2:example:condition}
The experimental condition is the same as Section~\ref{sec:experiment1:condition}, except for the manner in which the place's name is provided.

\begin{enumerate}[(a)]
 \item \textbf{Moving to the typical place from among several candidates}\\
    We demonstrate that the robot can move to a more typical and familiar place rather than a closer place. 
    In Section~\ref{sec:experiment1}, each bedroom was given the same number of training word data; three bedrooms were usually used about the same amount in that environment.
    However, it is also possible that the three bedrooms are not used to the comparable amount, i.e., each place's number of observed words in training data is unequal.
    For example, there may be one or two out of three bedrooms that are rarely used because they are used as guest rooms.

 \item \textbf{Multiple names associated with a same place}\\ 
    We demonstrated that the robot could estimate the path not only when multiple places with the same name existed, but also when multiple names were assigned to the same place.
    For example, we consider that a living room is represented by three names, including local names in this home environment.
    The robot is required to move to the same place irrespective of which name is chosen.

 \item \textbf{Conjunction relationship in words representing places}\\ 
    When two place names are provided in a sentence, 
    the emission probabilities represent the conjunction relationship between all places indicated by two names if their parts are covered or shared.
    In other words, it brings a similar effect on the {\it AND} operation between two places.
    For example, the user can instruct the robot in more detail; `{\it Go to the north bedroom.}' rather than `{\it bedroom}'.

 \item \textbf{Disjunction relationship in words representing places}\\ 
    When two place names are provided in a sentence, 
    the emission probabilities represent the disjunction relationship between places indicated by two names if they are located separately.
    In other words, it causes a similar effect on the {\it OR} operation between two places.
    For example, the user can instruct the robot with `{\it Go to kitchen or dining-room.}', when the user wants the robot to move to any one of the two places.
    The words given by the user in training are the same as Section~\ref{sec:experiment1:condition}. 
\end{enumerate}

\subsection{Result}
\label{sec:experiment1-2:example:result}

\begin{figure}[tb]
  \begin{center}
    \begin{tabular}{c}
    \includegraphics[clip, width=0.94\hsize]{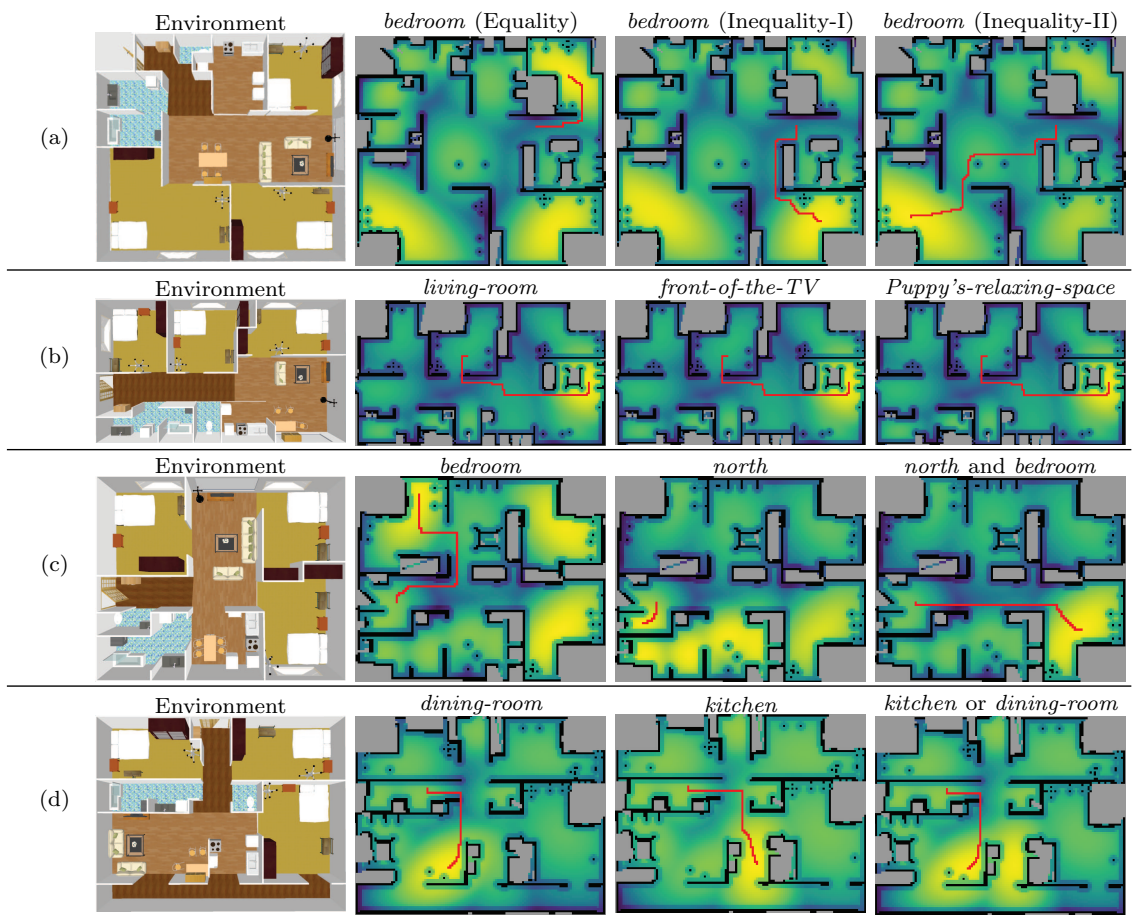}

    \end{tabular}
    \caption{Effective examples of estimated results by SpCoNavi. Environments and results of emission probabilities (color maps on log scale) and path-trajectories (red) for each place name. Examples in each case are shown as follows: (a) Moving to the typical place among several candidates; (b) Multiple names associated with a place; (c) Conjunction relationship between two place names; (d) Disjunction relationship between two place names.}
    \label{fig:result_SIGVerse_example}
  \end{center}
\end{figure}

Figure~\ref{fig:result_SIGVerse_example} (a) shows the estimated result of the case of moving to the typical place among several candidates, when the target place was `{\it dining-room}'.
In equality, three bedrooms were given as a comparable amount.
In inequality-I, the upper-right bedroom was less used than the other bedrooms.
In inequality-II, the lower-left bedroom was more used than the other bedrooms.
Consequently, the equality estimated the path that led to the closest place among the three bedrooms, 
inequality-I estimated a path that led to the closer place between two typical bedrooms, rather than the closest place, 
and inequality-II estimated a path that led to the farthest place.
If a nearby instructed place is rarely used, it implies a lower log-likelihood, and if a far instructed place is often used, it implies a higher log-likelihood.
Therefore, the robot can navigate to a suitable place. 

Figure~\ref{fig:result_SIGVerse_example} (b) shows the estimated result of the case of the multiple names associated with a place, when target places were `{\it living-room}', `{\it front-of-the-TV}', and `{\it Puppy's-relaxing-space}', respectively.
Consequently, the robot could estimate the same path with the probabilities for each word representing the same place.
The spatial concepts including local names that learned in this environment could be applied to the path planning by our method.

Figure~\ref{fig:result_SIGVerse_example} (c) shows the estimated result of the conjunction relationship in words representing places, when target places were `{\it bedroom}', `{\it north}', and `{\it north} and {\it bedroom}', respectively.
In the \textit{bedroom}, the result was the same as Section~\ref{sec:experiment1:result}.
In the \textit{north}, the robot learned the area of the lower side in this environment as `{\it north}'.
Places on the lower side showed higher emission probabilities than those on the upper side.
In {\it north} and {\it bedroom}, the robot could estimate the path to the place that matched both areas indicated by two words.

Figure~\ref{fig:result_SIGVerse_example} (d) shows the estimated result of the disjunction relationship in words representing places, when target places were `{\it kitchen}', `{\it dining-room}', and `{\it kitchen} or {\it dining-room}', respectively.
The results showed the path to a target place when the robot was given one word in the {\it dining-room} and the {\it kitchen}.
Moreover, when the robot was given two words representing different places, emission probabilities of both places showed higher values than others.

Although limited to some cases, our method could properly estimate the emission probabilities representing intended places without syntactic analysis in the sentence; the robot can automatically estimate the destination area from the relationship between places and names without having the meaning of {\it OR} or {\it AND} separately.
We believe that in the future work, we will combine it with parsing and semantic analysis to achieve a more complex and comprehensive linguistic navigation.

\section{Experiment II: Real robot environment}
\label{sec:experiment2}
Here, we demonstrate qualitatively that the robot, after acquiring the spatial concepts, can plan appropriate paths using our method in a real environment.

\subsection{Condition}
\label{sec:experiment2:condition}
The user says /** {ni iqte}/ in Japanese (which means `{\it Go to} **.' in English) as the speech instruction. 
{**} denotes the phoneme sequence for each place name.
The experimental environment was identical to that in the open dataset {\tt albert-b-laser-vision}{\footnote{{Dataset is available in \url{https://dspace.mit.edu/handle/1721.1/62291}}}}, which was obtained from the robotics dataset repository (Radish)~\cite{Radish}. 
{This dataset is a log file that contains the odometry, laser range data, and image data. 
The details of the properties of the robot are described in the dataset.}
Further, we used the global parameters $\Theta_{G}$ (map, parameters of spatial concept, and language model) of the maximum likelihood particle at step 50 learned in SpCoSLAM 2.0~\cite{ataniguchi2020spcoslam2}.
We used the {\tt navigation} and {\tt costmap\_2d} package in the ROS to obtain a global cost map. 
The planning horizon was $T=300$, and the $N$-best speech recognition was $N=10$.
Julius dictation-kit-v4.4 (DNN-HMM decoding)~\cite{lee2009recent} was used for speech recognition.
The microphone was a SHURE PG27-USB.
The robot's initial position was set at the center of the map. 
Here, the state $x_{t}$ is expressed discretely for each movable cell in the occupancy grid map $m$. 
The action $u_{t}$ is assumed to move one cell from the current position on the map per time-step by being discretized into forward, backward, side-to-side, and diagonal movements.

\subsection{Result}
\label{sec:experiment2:result}

\begin{figure}[tb]
  \begin{center}
    \begin{tabular}{c}
    \includegraphics[clip, width=0.96\hsize]{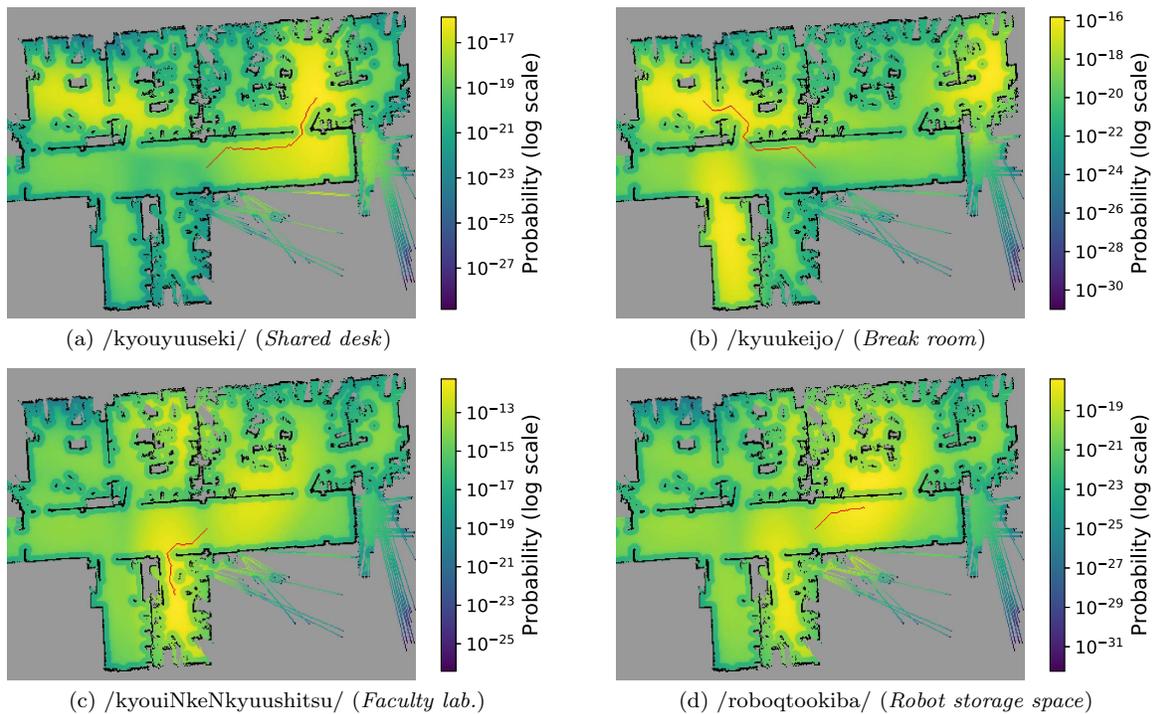}    
    

      
    \end{tabular}
    \caption{
        Emission probability for each cell on the map (color map on log scale) and path-trajectory (red) for each speech instruction in the real environment.
        Places instructed by the utterances were as follows: 
        (a) right side of the large room in the upper-right, (b) narrow room on the upper-right edge, and the room of the upper-left edge, (c) lower-right room, and (d) left side of the large room in the upper-right.
    }
    \label{fig:result}
  \end{center}
\end{figure}

Figure~\ref{fig:result} shows the results for path planning.
The log-likelihood values of the place that was instructed by the user's speech instruction were higher than other places.
This means that the reward value indicating the destination was changed by the speech instruction.
Consequently, the trajectory toward the instructed place was estimated in a real environment.

{
The error case exposed in the actual environment experiment is shown in Fig.~\ref{fig:result}~(d). In this case, the trajectory did not reach the target place but stopped in front of the wall, although the actual target place was behind the wall of the stop position. This case occurred owing to the fact that the range of the learned spatial concept exceeded the wall. This issue occurs because the position distribution of the spatial concept is represented by a Gaussian distribution; however, it is solved via application of an approach that deals with the shape of the room based on the conditional random field in~\cite{katsumata2019spcomapping}.
}

The NSR average value for nine taught places in our method was 0.667, the same value as in the conventional method (C). 
A navigation result depends on learning result of the spatial concepts and the speech recognition result based on the learned word dictionary.
It is suggested that more accurate global parameter's learning results will bring more accurate navigation.
However, because generalization, i.e., segmentation and clustering for places and speeches, can suppress data noise and uncertainty, we consider that our method works sufficiently well in many cases, even if a slight mistake at the learning exists.
For example, in Fig.~\ref{fig:result}~(c), the correct phoneme sequence spoken was /kyouiNkeNkyuushitsu/ (\textit{Faculty lab.} in English).
However, the robot succeeded in speech recognition of the word /kyoiNkeNkiyushitsu/, and it could move to the correct target place.

\section{Conclusion}
\label{sec:conclusion}

We proposed SpCoNavi\footnote{Source code is available in \url{https://github.com/a-taniguchi/SpCoNavi.git}.}: A path planning method with only speech instruction as an input that uses spatial concepts acquired autonomously by the robot.
Unlike many conventional path planning methods, our method need not specify goal-points on the map, and navigates to the destination based on speech interaction with the user. 
Additionally, we discussed the relationship between RL and our method based on CaI, and the approximate inference of SpCoNavi based on the heuristic extension of the conventional method.
Experimental results showed that our method estimated the trajectory to a place instructed by a speech instruction.

Our method is based on global path planning in a static environment map.
When the robot operates, it is assumed that it moves while performing self-localization via MCL along the estimated path. 
If the robot receives a new speech instruction during navigation, it will be able to move to a new destination by recalculating a trajectory.
Then, if a position shift occurs or an obstacle appears, local path planning will be combined to overcome these issues. A theoretical extension of SpCoNavi to deal with dynamic environments will be dealt with in a future study.

Our method has the advantage that it can be also applied to different POMDP-based models for spatial concepts, e.g.,~\cite{isobe2017learning,katsumata2019spcomapping}.
SpCoNavi was assumed to use parameters estimated by SpCoSLAM.
By converting semantic maps rather than spatial concepts into random variable formats compatible with SpCoNavi, the applicability of our approach is expected to increase, e.g., extracting semantic information from guide maps in buildings.
Moreover, because spatial concepts do not depend on mobile robot mechanisms, it will be possible to transfer spatial concepts from different robots.

SpCoNavi is a method for probabilistic inference concerning decision-making on a Bayesian generative model of SpCoSLAM. 
In other words, decision-making was formulated by estimating the probabilistic distribution of path-trajectories.
The theoretical framework of the CaI is associated with stochastic optimal and model predictive control, as well as RL~\cite{levine2018reinforcement, rawlik2013stochastic}.
We believe that this framework will be highly applicable and have high generality.

We consider that the navigation approach from speech information adopted in our method can also be applied to visual images.  
In other words, when showing an image of the target place to the robot, it can move to that place.
Moreover, the robot can navigate from images it has never seen before based on features similarities.

In a future study, we will attempt to reduce the computational complexity for real-time navigation further by creating a topological map or combining it with other planning algorithms~\cite{svestka1996probabilistic, lavalle1998rapidly}.

As a future prospect, a theoretical framework for probabilistic inference that integrates exploration by active-learning-based SpCoSLAM with the exploitation by SpCoNavi will be constructed. 
{Dealing with instructions including an unknown place and out-of-vocabulary words is an important topic that will be explored in the future.}
The robot will become capable of making decisions by switching navigation and acquisition of spatial concepts as required without possessing prior knowledge of the environment. 
This leads to further improvement of robot autonomy.


\section*{Acknowledgments}
This work was supported by the JST AIP-PRISM, under Grant JPMJCR18ZC; JST CREST, under Grant JPMJCR15E3 including AIP challenge program, Japan; and JSPS KAKENHI, under Grant JP20K19900, JP17J07842, JP16H06561, and JP16K12497.

The authors thank Cyrill Stachniss for providing {\tt albert-b-laser-vision} dataset. 
The authors also thank Kazuya Asada and Keishiro Taguchi for providing virtual home environments and training datasets for spatial concepts in SIGVerse simulator.


\appendix
\section{Formulation of the generative process of SpCoSLAM and SpCoNavi}
\label{apdx:SpCoSLAM:overview:formulation}
The details of the formulation of the generative process represented by the graphical model of SpCoSLAM and SpCoNavi are described as follows:
\begin{eqnarray}
\pi       &\sim& {\rm DP}(\alpha ) \label{eq:seisei1} \\
\phi_{l}  &\sim& {\rm DP}(\gamma ) \label{eq:seisei5} \\
\theta_{l}&\sim& {\rm Dir}(\chi) \label{eq3:seisei3} \\
W_{l}     &\sim& {\rm Dir}(\beta) \label{eq:seisei3} \\
LM        &\sim&  p(LM \mid \lambda) \label{eq2:seiseilm} \\
\Sigma_{k}&\sim& {\cal IW}( V_{0}, \nu _{0} )  \label{eq:seisei7} \\
\mu_{k}   &\sim& {\cal N}( m_{0}, \Sigma_{k} / \kappa _{0} ) \label{eq:seisei8} \\
x_{t}     &\sim& p(x_{t} \mid x_{t-1},u_{t}) \label{eq:seisei9} \\
z_{t}     &\sim& p(z_{t} \mid x_{t},m) \label{eq:seisei10} \\
C_{t}     &\sim& {\rm Mult}(\pi) \label{eq:seisei2} \\
i_{t}     &\sim& p(i_{t} \mid x_{t}, \mbox{\boldmath{$\mu $}}, \mbox{\boldmath{$\Sigma$}}, \mbox{\boldmath{$\phi$}}, C_{t}) \label{eq:seisei6} \\
f_{t}     &\sim& {\rm Mult}(\theta_{C_{t}}) \label{eq3:seisei2} \\
S_{t}     &\sim& p(S_{t} \mid {\bf{W}}, C_{t},LM) \label{eq2:seisei4b} \\
y_{t}     &\sim& p(y_{t} \mid S_{t},AM) \label{eq2:seiseiYt} 
\label{eq:seisei}
\end{eqnarray}
where ${\rm DP()}$ represents the Dirichlet process, ${\rm Dir()}$ is Dirichlet distribution, ${\cal IW()}$ is inverse-Wishart distribution, ${\cal N}()$ is multivariate normal distribution, and ${\rm Mult()}$ is multinomial distribution.
See \cite{murphy2012machine} for the specific formulas of the above probability distributions.

The probability distribution of Equation~(\ref{eq:seisei9}) represents a motion model---a state transition model---in SLAM.
The probability distribution of Equation~(\ref{eq:seisei10}) represents a measurement model in SLAM.

The probability distribution of Equation~(\ref{eq:seisei6}) can be defined as 
\begin{eqnarray}
p(i_{t} \mid x_{t},\mbox{\boldmath{$\mu $}},\mbox{\boldmath{$\Sigma$}} ,\mbox{\boldmath{$\phi$}}, C_{t})
&=& \cfrac{{\cal N}(x_{t} \mid \mu _{i_{t}}, \Sigma_{i_{t}}){\rm Mult}(i_{t} \mid \phi_{C_{t}})}{\sum_{i_{t}=j} {\cal N}(x_{t} \mid \mu _{j}, \Sigma_{j}){\rm Mult}(j \mid \phi_{C_{t}})}~.
\label{eq:it}
\end{eqnarray}

The probability distribution of Equation~(\ref{eq2:seisei4b}) is approximated by unigram rescaling~\cite{gildea1999topic}, as 
\begin{eqnarray}
p(S_{t} \mid {\bf{W}}, C_{t},LM)&{\approx }& p(S_{t} \mid LM) \prod_{B_{t}} \frac{{\rm Mult}(S_{t,b} \mid W_{C_{t}})}{\sum_{c'}{\rm Mult}(S_{t,b} \mid W_{c'})},
\label{eq2:st_UR}
\end{eqnarray}
where $B_{t}$ denotes the number of words in the sentence and $S_{t,b}$ is $b$-th word in the sentence at time-step $t$.

\section{Learning procedure of SpCoSLAM for each step}
\label{apdx:SpCoSLAM:learning:procedure}
The learning procedure of SpCoSLAM for each step is described as follows:
\begin{enumerate}[(a)]
\item The robot obtains weighted finite-state transducer (WFST) speech recognition results $\mathcal{L}_{1:t}$ from the user's speech signals $y_{1:t}$ using a language model $LM$. 
The WFST is a word graph, i.e., a lattice format, which represents the $N$-best speech recognition results alternatively.
In the initialization, a phoneme dictionary is provided as the language model $LM$ without a prior word list.
\item The WFST speech recognition results $\mathcal{L}_{1:t}$ are segmented to the word sequences $S_{1:t}$, using an unsupervised word segmentation approach called latticelm~\cite{neubig2012bayesian}. 
\item The latent variable $x_{t}$ and importance weight $\omega_{z}$ regarding self-localization are obtained by the grid-based FastSLAM 2.0 {from control data $u_{t}$, depth data $z_{t}$ and particles that represent the self-positions $x_{t-1}$ at the previous time}.
\item The latent variables $ i_{t}, C_{t} $ of spatial concepts are sampled by the proposal distribution on the particle filter.
\item The importance weights $\omega_{s}$, $\omega_{f}$ are obtained as the marginal likelihoods of observations $ S_{t}, f_{t}$.
\item The environmental map $m$ is updated by self-positions $x_{0:t}$ and depth data $z_{1:t}$.
\item The set of model parameters $\Theta$ of the spatial concepts are estimated from the observation $f_{1:t}$ and sampled variables $ x_{0:t}, \mathbf{C}_{1:t}$. 
\item The language model $LM$ is updated by adding words $S_{1:t}^{\ast}$ in a particle of maximum weight to the initial dictionary.
\item The particles are resampled according to their weights $\omega_{t}= \omega_{z} \cdot \omega_{s} \cdot \omega_{f}$. 
\end{enumerate}
Steps (b) -- (g) are performed for each particle.
See the original paper~\cite{ataniguchi_IROS2017} for details.

\section{RL and control as probabilistic inference}
\label{apdx:SpCoNavi:CaI}

The theoretical gap between the control problems including RL and the probabilistic inference in the generative model was bridged by CaI~\cite{levine2018reinforcement}.

\begin{figure}
    \begin{center}
        \includegraphics[width=0.7\textwidth]{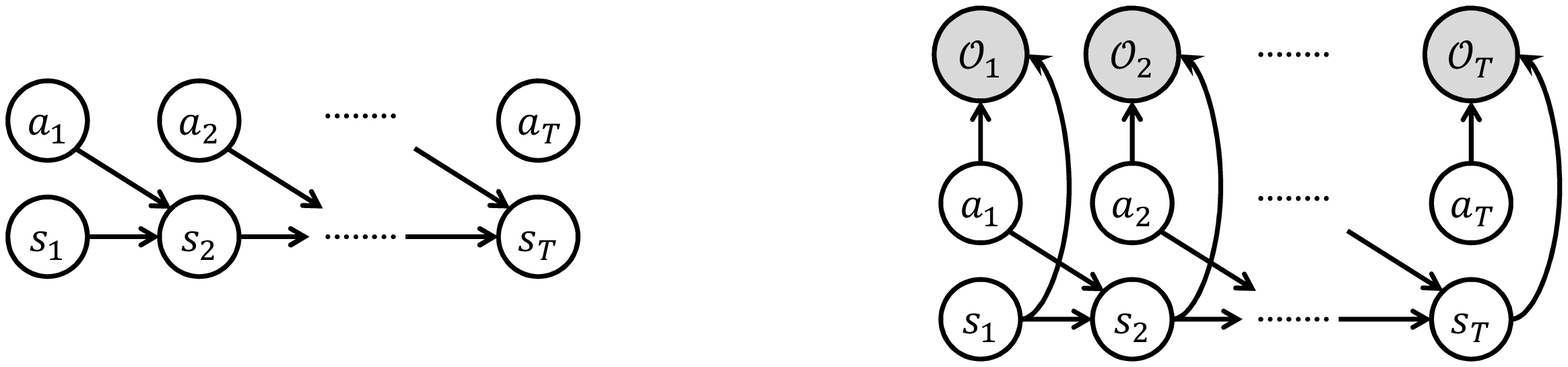}
        \caption{
            Left: Graphical model of MDP with states and actions. 
            Right: Graphical model for CaI with the optimality variables. 
            This additional variable is a binary random variable, where ${\cal{O}}_{t}=1$ denotes that time-step $t$ is optimal.
        }
        \label{fig:CaI}
    \end{center}
\end{figure}

In general decision-making problems including RL, a policy that maximizes the expected cumulative reward is estimated as 
\begin{eqnarray}
\vartheta^{\star} &=& \argmax_{\vartheta} \sum_{t=1}^{T}  \mathbb{E}_{(s_{t}, a_{t}) \sim p(s_{t},a_{t} \mid \vartheta)} \left[ r(s_{t}, a_{t}) \right],
\label{eq:expected_reward}
\end{eqnarray}
where $r(s_{t}, a_{t})$ is a reward function, $s_{t}$ is a state variable, $a_{t}$ is an action variable, $\vartheta$ is a parameter for the policy function, and $\vartheta^{\ast}$ is an optimal policy parameter.
Note that $s_{t}$ and $a_{t}$ correspond to $x_{t}$ and $u_{t}$ in our method, respectively.

From the viewpoint of CaI, the planning problem can be formulated as an inference on the probabilistic graphical model. 
Figure~\ref{fig:CaI} shows the graphical models of Markov decision process (MDP) with an optimality variable ${\cal{O}}_{t}$. 
In the graphical model for CaI, the distribution denoting the generative process on the binary random variable ${\cal{O}}_{t}$ is represented as 
\begin{eqnarray}
p({\cal{O}}_{t}=1 \mid s_{t}, a_{t}) &=& \exp(r(s_{t}, a_{t})).
\label{eq:bainary}
\end{eqnarray}

The maximum a posteriori inference in the posterior distribution $p(\tau \mid o_{1:T})$ corresponds to a type of planning problems. 
Here, trajectory is $\tau=\{ s_{1:T}, a_{1:T} \}$ and the set of optimality variables is $o_{1:T} = \{ {\cal{O}}_{t}=1 \}^{T}_{t=1}$.
The posterior distribution over actions when we condition ${\cal{O}}_{t}=1$ for all $t \in \{ 1, \dots, T\}$ is shown as 
\begin{eqnarray}
p(\tau \mid o_{1:T})
&\propto& p(s_{1})\prod_{t=1}^{T} p(s_{t+1} \mid s_{t}, a_{t}) p({\cal{O}}_{t}=1 \mid s_{t}, a_{t}) \nonumber \\
&=& \underbrace{ \left[ p(s_{1})\prod_{t=1}^{T} p(s_{t+1} \mid s_{t}, a_{t}) \right] }_\text{State-transition with action} \exp \underbrace{ \left( \sum_{t=1}^{T} r(s_{t}, a_{t}) 
\right) }_\text{Cumulative reward}. 
\label{eq:posterior}
\end{eqnarray}
It means that the optimalities $o_{1:T}$ are given as observations in a HMM-style models.
Therefore, the trajectory probability is given by the product between its probability to occur according to the dynamics and the exponential of the cumulative reward along that trajectory.

In this case, the policy function can be shown as $\pi_{\vartheta}(s_{t}, a_{t}) = p(a_{t} \mid s_{t}, \vartheta)$.
The optimal policy function can be shown as $p(a_{t} \mid s_{t}, \vartheta^{\star}) \approx p(a_{t} \mid s_{t}, {o}_{t:T})$; then, the right side is not related to the parameter $\vartheta$.

Levine have described that the CaI makes it possible to apply various techniques of probabilistic inference, e.g., forward-backward algorithm and variational inference, to control and planning problems.
See the paper~\cite{levine2018reinforcement} for details.


\bibliographystyle{tADR}
\bibliography{./ForJabReff_utf8}  

\begin{thebibliography}{10}
\providecommand{\url}[1]{\normalfont{#1}}
\providecommand{\urlprefix}{Available from: }
\providecommand{\eprint}[2][]{\url{#2}}

\bibitem{taniguchi2015symbol}
Taniguchi T, Nagai T, Nakamura T, Iwahashi N, Ogata T, Asoh H. Symbol emergence
  in robotics: A survey. Advanced Robotics. 2016;\hspace{0pt}30:706--728.

\bibitem{taniguchi2019langrobo}
Tangiuchi T, Mochihashi D, Nagai T, Uchida S, Inoue N, Kobayashi I, Nakamura T,
  Hagiwara Y, Iwahashi N, Inamura T. Survey on frontiers of language and
  robotics. Advanced Robotics. 2019;\hspace{0pt}33(15-16):700--730.
  \eprint{https://doi.org/10.1080/01691864.2019.1632223}.
  \urlprefix\url{https://doi.org/10.1080/01691864.2019.1632223}.

\bibitem{kostavelis2015semantic}
Kostavelis I, Gasteratos A. Semantic mapping for mobile robotics tasks: A
  survey. Robotics and Autonomous Systems. 2015;\hspace{0pt}66:86--103.

\bibitem{landsiedel2017review}
Landsiedel C, Rieser V, Walter M, Wollherr D. A review of spatial reasoning and
  interaction for real-world robotics. Advanced Robotics.
  2017;\hspace{0pt}31(5):222--242.

\bibitem{ataniguchi_IROS2017}
Taniguchi A, Hagiwara Y, Taniguchi T, Inamura T. Online spatial concept and
  lexical acquisition with simultaneous localization and mapping. In:
  {Proceedings of the IEEE/RSJ International Conference on Intelligent Robots
  and Systems (IROS)}. 2017. p. 811--818.

\bibitem{ataniguchi2020spcoslam2}
Taniguchi A, Hagiwara Y, Taniguchi T, Inamura T. Improved and scalable online
  learning of spatial concepts and language models with mapping. Autonomous
  Robots. 2020;\hspace{0pt}44(6):927--946.

\bibitem{toussaint2009robot}
Toussaint M. Robot trajectory optimization using approximate inference. In:
  {Proceedings of the International Conference on Machine Learning (ICML)}.
  ACM. 2009. p. 1049--1056.

\bibitem{levine2018reinforcement}
Levine S. Reinforcement learning and control as probabilistic inference:
  Tutorial and review. arXiv preprint arXiv:180500909. 2018;\hspace{0pt}.

\bibitem{sutton1998reinforcement}
Sutton RS, Barto AG. Reinforcement learning: An introduction. MIT press
  Cambridge. 1998.

\bibitem{mukadam2017simultaneous}
Mukadam M, Dong J, Dellaert F, Boots B. Simultaneous trajectory estimation and
  planning via probabilistic inference. In: Robotics: Science and systems.
  2017.

\bibitem{mukadam2019steap}
Mukadam M, Dong J, Dellaert F, Boots B. Steap: simultaneous trajectory
  estimation and planning. Autonomous Robots. 2019
  Feb;\hspace{0pt}43(2):415--434.

\bibitem{kretzschmar2016socially}
Kretzschmar H, Spies M, Sprunk C, Burgard W. Socially compliant mobile robot
  navigation via inverse reinforcement learning. The International Journal of
  Robotics Research. 2016;\hspace{0pt}35(11):1289--1307.

\bibitem{kinose2020integration}
Kinose A, Taniguchi T. Integration of imitation learning using {GAIL} and
  reinforcement learning using task-achievement rewards via probabilistic
  graphical model. Advanced Robotics. 2020;\hspace{0pt}:1--13.

\bibitem{svestka1996probabilistic}
Svestka P, Latombe J, Overmars~Kavraki L. Probabilistic roadmaps for path
  planning in high-dimensional configuration spaces. IEEE Transactions on
  Robotics and Automation. 1996;\hspace{0pt}12(4):566--580.

\bibitem{lavalle1998rapidly}
LaValle SM. Rapidly-exploring random trees: A new tool for path planning.
  Research Report 9811. 1998;\hspace{0pt}98(11).

\bibitem{karaman2011sampling}
Karaman S, Frazzoli E. Sampling-based algorithms for optimal motion planning.
  The international journal of robotics research.
  2011;\hspace{0pt}30(7):846--894.

\bibitem{kollar2010toward}
Kollar T, Tellex S, Roy D, Roy N. Toward understanding natural language
  directions. In: {Proceedings of the 5th ACM/IEEE International Conference on
  Human-Robot Interaction (HRI)}. IEEE. 2010. p. 259--266.

\bibitem{nair2018visual}
Nair AV, Pong V, Dalal M, Bahl S, Lin S, Levine S. Visual reinforcement
  learning with imagined goals. In: {Proceedings of the Advances in Neural
  Information Processing Systems (NeurIPS)}. 2018. p. 9191--9200.

\bibitem{fu2018from}
Fu J, Korattikara A, Levine S, Guadarrama S. From language to goals: Inverse
  reinforcement learning for vision-based instruction following. In:
  {Proceedings of the International Conference on Learning Representations
  (ICLR)}. 2019.

\bibitem{anderson2018vision}
Anderson P, Wu Q, Teney D, Bruce J, Johnson M, S{\"u}nderhauf N, Reid I, Gould
  S, van~den Hengel A. Vision-and-language navigation: Interpreting
  visually-grounded navigation instructions in real environments. In:
  {Proceedings of the IEEE Conference on Computer Vision and Pattern
  Recognition (CVPR)}. 2018. p. 3674--3683.

\bibitem{noguchi2018navigation}
Noguchi W, Iizuka H, Yamamoto M. Navigation behavior based on self-organized
  spatial representation in hierarchical recurrent neural network. Advanced
  Robotics. 2019;\hspace{0pt}33(11):539--549.
  \eprint{https://doi.org/10.1080/01691864.2019.1566088}.
  \urlprefix\url{https://doi.org/10.1080/01691864.2019.1566088}.

\bibitem{fang2019smt}
Fang K, Toshev A, Fei-Fei L, Savarese S. Scene memory transformer for embodied
  agents in long-horizon tasks. In: {Proceedings of the IEEE Conference on
  Computer Vision and Pattern Recognition (CVPR)}. 2019.

\bibitem{lotfi2018WRS}
El~Hafi L, Isobe S, Tabuchi Y, Katsumata Y, Nakamura H, Fukui T, Matsuo T,
  Ricardez G, Yamamoto M, Taniguchi A, Hagiwara Y, Taniguchi T. System for
  augmented human-robot interaction through mixed reality and robot training by
  non-experts in customer service environments. Advanced Robotics.
  2020;\hspace{0pt}34(3-4):157--172.

\bibitem{thrun2005probabilistic}
Thrun S, Burgard W, Fox D. Probabilistic robotics. MIT Press. 2005.

\bibitem{aldous1985exchangeability}
Aldous D. Exchangeability and related topics. {\'E}cole d'{\'E}t{\'e} de
  Probabilit{\'e}s de Saint{-}Flour XIII{-}1983. 1985;\hspace{0pt}:1--198.

\bibitem{doucet2000rao}
Doucet A, De~Freitas N, Murphy K, Russell S. {R}ao-{B}lackwellised particle
  filtering for dynamic {B}ayesian networks. In: Proceedings of the 16th
  conference on uncertainty in artificial intelligence. Morgan Kaufmann
  Publishers Inc.. 2000. p. 176--183.

\bibitem{montemerlo2003fastslam}
Montemerlo M, Thrun S, Koller D, Wegbreit B, et~al.. {FastSLAM} 2.0: An
  improved particle filtering algorithm for simultaneous localization and
  mapping that provably converges. In: {Proceedings of the International Joint
  Conference on Artificial Intelligence (IJCAI)}. 2003. p. 1151--1156.

\bibitem{gridbasedfastslam2007}
Grisetti G, Stachniss C, Burgard W. Improved techniques for grid mapping with
  {R}ao-{B}lackwellized particle filters. IEEE Transactions on Robotics.
  2007;\hspace{0pt}23:34--46.

\bibitem{taniguchi_spcoa}
Taniguchi A, Taniguchi T, Inamura T. Spatial concept acquisition for a mobile
  robot that integrates self{-}localization and unsupervised word discovery
  from spoken sentences. IEEE Transactions on Cognitive and Developmental
  Systems. 2016;\hspace{0pt}8(4):285--297.

\bibitem{dellaert1999monte}
Dellaert F, Fox D, Burgard W, Thrun S. Monte carlo localization for mobile
  robots. In: {Proceedings of the IEEE International Conference on Robotics and
  Automation (ICRA)}. Vol.~2. IEEE. 1999. p. 1322--1328.

\bibitem{inamura2020sigverse}
Inamura T, Mizuchi Y. Sigverse: A cloud-based vr platform for research on
  social and embodied human-robot interaction. arXiv preprint arXiv:200500825.
  2020;\hspace{0pt}.

\bibitem{ros}
Quigley M, Conley K, Gerkey BP, Faust J, Foote T, Leibs J, Wheeler R, Ng AY.
  Ros: an open-source robot operating system. In: Proceedings of the icra
  workshop on open source software. 2009.

\bibitem{HSR2019}
Yamamoto T, Terada K, Ochiai A, Saito F, Asahara Y, Murase K. Development of
  human support robot as the research platform of a domestic mobile
  manipulator. ROBOMECH Journal. 2019 Apr;\hspace{0pt}6(1):4.
  \urlprefix\url{https://doi.org/10.1186/s40648-019-0132-3}.

\bibitem{Radish}
Howard A, Roy N. Radish: Robotics research datasets. 2003.
  \urlprefix\url{https://dspace.mit.edu/handle/1721.1/62236}.

\bibitem{lee2009recent}
Lee A, Kawahara T. Recent development of open-source speech recognition engine
  {J}ulius. In: Proceedings of the apsipa asc. 2009. p. 131--137.

\bibitem{katsumata2019spcomapping}
Katsumata Y, Taniguchi A, Hagiwara Y, Taniguchi T. Semantic mapping based on
  spatial concepts for grounding words related to places in daily environments.
  Frontiers in Robotics and AI. 2019;\hspace{0pt}6:31.

\bibitem{isobe2017learning}
Isobe S, Taniguchi A, Hagiwara Y, Taniguchi T. Learning relationships between
  objects and places by multimodal spatial concept with bag of objects. In:
  {Proceedings of the International Conference on Social Robotics (ICSR)}.
  Springer. 2017. p. 115--125.

\bibitem{rawlik2013stochastic}
Rawlik K, Toussaint M, Vijayakumar S. On stochastic optimal control and
  reinforcement learning by approximate inference. In: {Proceedings of the
  International Joint Conference on Artificial Intelligence (IJCAI)}. AAAI
  Press. 2013. p. 3052--3056.

\bibitem{murphy2012machine}
Murphy KP. Machine learning: a probabilistic perspective. Cambridge, MA: MIT
  Press. 2012.

\bibitem{gildea1999topic}
Gildea D, Hofmann T. Topic-based language models using em. In: In proceedings
  of the 6th european conference on speech communication and technology
  (eurospeech). 1999.

\bibitem{neubig2012bayesian}
Neubig G, Mimura M, Kawahara T. {B}ayesian learning of a language model from
  continuous speech. IEICE Transactions on Information and Systems.
  2012;\hspace{0pt}95(2):614--625.

\end{thebibliography}

\end{document}